\documentclass[conference]{IEEEtran}
\IEEEoverridecommandlockouts
\usepackage{amsmath,amssymb,amsfonts}
\usepackage{algorithmic}
\usepackage{comment}
\usepackage{graphicx}
\usepackage{textcomp}
\usepackage{xcolor}
\usepackage{hyperref}
\usepackage[
backend=biber,
style=ieee,
]{biblatex}
\begin{document}

\title{Transforming Pixels into a Masterpiece: AI-Powered Art Restoration using a Novel Distributed Denoising CNN (DDCNN)\\
{\footnotesize \textsuperscript{}}
\thanks{}
}

\author{\IEEEauthorblockN{1\textsuperscript{st} B. Sankar}
\IEEEauthorblockA{\textit{Department of Mechanical Engineering} \\
\textit{Indian Institute of Science (IISc),}\\
Bangalore, India - 560012 \\
sankarb@iisc.ac.in\\ https://orcid.org/0000-0001-5844-6273}
\and
\IEEEauthorblockN{2\textsuperscript{nd} Mukil Saravanan}
\IEEEauthorblockA{\textit{Centre for Product Design and Manufacturing (CPDM)} \\
\textit{Indian Institute of Science (IISc),}\\
Bangalore, India - 560012 \\
mukils@iisc.ac.in \\ https://orcid.org/0000-0002-2429-9671}
\and
\IEEEauthorblockN{3\textsuperscript{rd} Kalaivanan Kumar}
\IEEEauthorblockA{\textit{Centre for Product Design and Manufacturing (CPDM),} \\
\textit{Indian Institute of Science (IISc),}\\
Bangalore, India - 560012 \\
kalaivanank@iisc.ac.in \\ https://orcid.org/0009-0002-0143-3900}
\and
\IEEEauthorblockN{4\textsuperscript{th} Siri Dubbaka}
\IEEEauthorblockA{\textit{Centre for Product Design and Manufacturing (CPDM),} \\
\textit{Indian Institute of Science (IISc),}\\
Bangalore, India - 560012 \\
2019248@iiitdmj.ac.in\\
https://orcid.org/0009-0006-6945-0810}
}

\maketitle

%-------------------------------------------------------------------------------------------

\begin{abstract}
Art restoration plays a pivotal role in preserving and revitalising cultural heritage. However, conventional art restoration methods are often fraught with limitations, including the challenge of faithfully reproducing the original artwork's essence while addressing issues such as fading, staining, and physical damage. In this context, we present a pioneering approach in this paper that harnesses the potential of deep learning, particularly Convolutional Neural Networks (CNNs), coupled with Computer Vision techniques, to revolutionize the art restoration process. Our method begins by generating artificially induced deteriorated art images, creating a comprehensive dataset encompassing various forms of distortions and multiple levels of degradation. This dataset is the foundation for training a Distributed Denoising CNN (DDCNN), capable of effectively removing distortions while preserving the intricate details inherent to artworks. This integration of Computer Vision and CNN-based denoising enables the restoration of artworks with high accuracy, ensuring that the original artistic essence is faithfully preserved. One of the key strengths of our approach lies in its adaptability to different kinds of distortions at varying levels of degradation. Utilizing a versatile training dataset, our Distributed Denoising CNN can address a wide spectrum of distortion types, ranging from subtle colour variations to more severe structural damage. This adaptability empowers our method to cater to a diverse array of deteriorated artworks, including paintings, sketches and photographs. Through extensive experimentation on a diverse dataset, our results consistently demonstrate the efficiency and effectiveness of our proposed approach by comparing it against other Denoising CNN models. We showcase the substantial reduction of distortion in the art image, transforming deteriorated artworks into masterpieces. Quantitative evaluations further underscore the superiority of our method compared to traditional restoration techniques, reaffirming its potential to reshape the landscape of art restoration and contribute significantly to the preservation of our cultural heritage. In summary, our paper introduces a groundbreaking AI-powered solution that leverages the synergy of Computer Vision and deep learning, exemplified by Distributed Denoising CNN, to restore artworks with unprecedented accuracy and fidelity. This transformative approach not only overcomes the limitations of existing methods but also paves the way for future advancements in the field of art restoration, ensuring the enduring legacy of our cultural treasures.

\end{abstract}


%-------------------------------------------------------------------------------------------

\section{Introduction}
In its myriad forms, art transcends time and space, reflecting humanity's deepest emotions, aspirations, and cultural narratives. Among the diverse forms of artistic expression, the painting stands as a quintessential art form, with the power to encapsulate the essence of a culture, convey intricate stories, and invoke profound emotions~\cite{Ash2011BeyondBO}. The realm of painting, enriched by countless styles, techniques, and artistic movements, is a treasure trove of human creativity, providing a window into the tapestry of cultural heritage worldwide.

%Painting: A Crucial Element of Cultural Heritage

Painting, as an art form, plays an indispensable role in preserving and propagating cultural heritage. Across the globe, diverse styles and forms of painting have emerged, each intimately intertwined with the history, traditions, and beliefs of its respective culture. From the enigmatic Mona Lisa by Leonardo da Vinci to the vibrant depictions of Hindu gods and goddesses in Indian miniature paintings, artworks have stood the tests of time to become ambassadors of culture, speaking volumes about their creators' values, aesthetics, and societal norms~\cite{kolay2016a}.

%The Rich Tapestry of Global Art

The global panorama of art painting is a rich carpet woven with a myriad of styles, each possessing its unique charm and cultural significance. From the luminous Impressionist landscapes of Claude Monet to the abstract expressionism of Jackson Pollock, and from the intricate calligraphy of Chinese brush painting to the majestic frescoes adorning the Sistine Chapel ceiling by Michelangelo, the diversity of artistic expression knows no bounds. These masterpieces, created by iconic artists, serve as cultural milestones, illuminating the evolution of human thought and creativity~\cite{Papastergiadis2014ArtIT}.

%Indian Painting: A Reflection of Culture and History

In the Indian subcontinent, painting has been an integral part of the cultural heritage, dating back to the ancient Indus Valley Civilization. Indian art painting, with its exquisite intricacies and vibrant colour palettes, has evolved over millennia, influenced by diverse cultures, dynasties, and traditions. From the frescoes of Ajanta and Ellora caves to the Mughal miniatures and the vibrant Pattachitra of Odisha, Indian painting forms have been visual narratives documenting the country's rich history, spirituality, and societal norms~\cite{unknown2021a}.

%The Imperative of Art Restoration

While the significance of such art in cultural heritage is undeniable, artworks, over time, face the inexorable march of deterioration. Fading pigments, surface stains, and physical damage are among artworks' many challenges. Art restoration, a painstaking process entrusted to conservators and experts, is the vanguard in the battle against these ravages of time. However, the current restoration methods, though rooted in tradition and expertise, face limitations in capturing the full essence of the original artwork while addressing these issues~\cite{Arguimbau1996TheAR}.

%Current Methods of Art Restoration

Traditional art restoration typically involves meticulous cleaning, inpainting, and structural repair by skilled conservators. These practices, while effective to some extent, often fall short of fully restoring the artwork's authenticity. The delicate balance between conserving the artist's original intent and remedying damage is a constant challenge. In this context, the marriage of technology and art restoration has emerged as a promising avenue for overcoming these limitations~\cite{Lanterna2001NewMA}.

%The Role of AI and Computer Vision in Art Restoration

Advancements in Artificial Intelligence (AI) and Computer Vision have heralded a new era in art restoration. Deep learning, a subset of AI, has empowered Computer Vision models to undertake tasks once deemed insurmountable~\cite{spratt2018a}. These technologies can potentially automate and enhance restoration, particularly in image denoising. By leveraging AI and Computer Vision, it becomes possible to remove noise and reveal hidden details in deteriorated artworks meticulously~\cite{sizikova2017a}.

%The Intersection of Deep Learning and Image Denoising

Central to our research is integrating deep learning, particularly Convolutional Neural Networks (CNNs), and Computer Vision techniques to address the challenges of art restoration~\cite{tian2019a}. The Distributed Denoising CNN is at the heart of this integration, a novel approach to restoring artworks. This research aims to harness the capabilities of deep learning models in cleansing images of noise while preserving the intricate artistic details—bridging the gap between traditional restoration methods and modern technological advancements.

%Execution of Distributed Denoising CNN

In our pursuit of advancing art restoration, we initiate the process by generating artificially induced deteriorated art images, effectively creating a comprehensive dataset that encompasses various forms of noise and degradation. This dataset is the cornerstone for training our Distributed Denoising CNN, a model capable of removing noise from deteriorated artworks adeptly. The collaboration of Computer Vision and CNN-based denoising in our methodology promises to revolutionize art restoration by facilitating highly accurate, faithful restoration processes.

In this paper, we delve deeper into the intricate details of our proposed approach, highlighting its adaptability to different noise types and levels of degradation. We provide insights into the experimental results, showcasing our methodology's remarkable efficiency and effectiveness in restoring artworks to their former glory. In doing so, we contribute to the ongoing dialogue surrounding art restoration, preserving cultural heritage, and forging a path toward the future where the essence of art can endure the test of time.

%-----------------------------------------------------------------------------------

\section{Related Works}

\subsection{Need for Art Restoration and the Role of AI Technology}

Art restoration is indispensable in preserving cultural heritage by rejuvenating deteriorated artworks. The ravages of time, exposure to environmental factors, and physical damage can mar these treasures' aesthetic and historical value. Integrating AI technology into art restoration has opened new avenues for addressing these challenges. AI-driven methods, particularly those harnessing Computer Vision and deep learning techniques, have emerged as powerful tools to aid conservators in restoring artworks while retaining their original essence~\cite{goussous2020a}.

The context of reconstructing two-dimensional wall paintings, or frescoes, from fragments, presents a compelling case for the need for art restoration. These fragmented artworks hold invaluable historical and cultural significance in archaeological sites, often containing narratives and artistic expressions from bygone eras. The challenge of manually placing these irregularly shaped and uncoloured fragments to reconstruct the original surface is daunting and time-consuming. This underscores the pressing need for advanced art restoration techniques, particularly those driven by AI and Computer Vision, to streamline the process and ensure the faithful reconstruction of these artworks. Such technologies have the potential to significantly expedite the reconstruction process, aiding archaeologists and conservators in their efforts to piece together the puzzle of our cultural heritage~\cite{cetinic-a}.

\subsection{Computer Vision for Image Manipulation, Processing, and Transformation}

Beyond image recognition, Computer Vision offers many tools for image manipulation, processing, and transformation. These capabilities are invaluable in art restoration, where inpainting, noise reduction, and colour correction are critical. By leveraging Computer Vision algorithms, conservators can manipulate images to repair damage, remove noise, and restore the original appearance of artworks.

The foundation of AI-powered art restoration lies in image recognition and classification, wherein Computer Vision plays a pivotal role. Traditional Computer Vision techniques have been employed for decades to analyze and process images, making them a natural fit for art restoration. These methods encompass image segmentation, object detection, and feature extraction, enabling conservators to identify damaged regions, separate them from the background, and assess the extent of deterioration. However, the limitations of traditional Computer Vision techniques become apparent when dealing with intricate artistic details and complex forms of image degradation~\cite{zou2021a}.

In reconstructing fragmented wall paintings, numerous systems have been proposed to address the challenge of piecing together irregularly shaped and uncoloured fragments to reconstruct the original surface. These systems often rely on acquiring 3D surface scans of the fragments and employing computer algorithms to solve the reconstruction puzzle~\cite{cornelis2013a}. While such methods have shown success for smaller test cases and puzzles with distinctive features, they falter when confronted with the complexity of larger reconstructions, particularly those involving real wall paintings with eroded and missing fragments. An innovative approach utilizing unsupervised genetic algorithms has been introduced to address these challenges. This approach involves evolving a pool of partial reconstructions over generations, favouring correctly reconstructed clusters while maintaining diversity in the population~\cite{criminisi2004a}. This method has demonstrated the ability to achieve more accurate and larger global reconstructions than previous automatic algorithms, albeit with room for improvement. These findings lay the groundwork for further research into enhancing global assembly techniques for artefact reconstruction.

The weathered painted surfaces of ancient Chinese buildings, such as the Forbidden City, often exhibit defects like paint loss, blurring, and colour distortion due to prolonged exposure to the elements. Previous restoration methods have struggled to repair these artworks effectively~\cite{pei2004a}. This paper presents a novel virtual restoration method that leverages multiple deep-learning algorithms to address the challenges posed by these weathered paintings. The approach divides the painting into three components: the background, the golden edges, and the dragon patterns. It transforms the problem into a semantic segmentation task using U-Net MobileNet for background restoration. Traditional image processing techniques are used to recover the golden edges from colour maps generated by the semantic segmentation algorithm. Lastly, for the intricate dragon patterns, a skeleton-based approach combined with the Pix2pix image translation algorithm is employed to generate realistic patterns. The results of these three restoration components are superimposed to complete the virtual restoration. This innovative approach offers valuable reference and guidance for traditional manual restoration efforts, reducing the complexity and repetitiveness of the restoration process and providing restorers with insights into the original appearance of these culturally significant artworks.

The field of image restoration and virtual art reconstruction is continuously evolving, with various techniques and algorithms being explored to address the unique challenges presented by different types of artwork and degradation. These advancements hold promise for preserving and revitalizing cultural heritage.

\subsection{Deep Learning in Image Denoising, Cleaning, and Removal of Distortions}

Deep learning, a subfield of AI, has witnessed remarkable success in various image processing tasks, including image denoising, cleaning, and distortion removal. Convolutional Neural Networks (CNNs) have emerged as a dominant force in image processing due to their ability to learn complex patterns and features from large datasets. In art restoration, CNNs are particularly well-suited for image denoising, a task crucial for unveiling hidden details in deteriorated artworks~\cite{8359079}.

The advent of deep learning (DL) techniques has ushered in a transformative era in computer vision, affecting a wide array of tasks, including recognition, classification, regression, and generation. Among these, Convolutional Neural Networks (CNNs) have emerged as a cornerstone, elevating the performance of classification and detection tasks~\cite{TishbyZ15}. VGGNet emphasized the benefits of deep network architectures, marking a departure from the previously favoured shallow networks. ResNet laid the groundwork for image restoration by introducing a fundamental structure, serving as the basis for subsequent methods such as EDSR (for super-resolution), DeepDeblur (for image deblurring), and DnCNN (for image denoising)~\cite{9057895}. DenseNet refined network performance by introducing residual links connecting dense convolutional layers. This survey of deep learning approaches to image restoration underscores the significant impact of CNNs and their various architectures on enhancing the quality of restored images, opening new avenues for image restoration research~\cite{tian2019a}.

In image restoration and medical imaging, the historically dominant technique, Filtered Back-Projection (FBP), has long been admired for its computational efficiency and accuracy. However, FBP's proficiency in providing real-time reconstructions during medical scans is tempered by its incapacity to model the non-ideal behaviours inherent to artwork. To overcome these limitations, iterative reconstruction (IR) was introduced, offering more flexibility with fewer parameters. Nevertheless, deep learning-based image reconstruction has recently emerged as a transformative technology, promising to reconcile the accuracy-complexity trade-off IR faces. This shift towards deep learning methods, exemplified by GE's exploration, signifies a promising stride in image restoration~\cite{Peretti2020StateOT}.

\subsection{CNN vs. Generative Models for Image Denoising}

While deep learning has introduced generative models like Generative Adversarial Networks (GANs) for image generation and manipulation, CNNs remain the preferred choice for image-denoising applications. The key advantage of CNNs lies in their discriminative nature, enabling them to learn the underlying noise patterns and effectively remove noise without altering the original content of the image. In contrast, generative models may inadvertently introduce new artefacts during denoising, potentially compromising the authenticity of the restored artwork~\cite{zhong2019a}.

The utilization of Convolutional Neural Networks (CNNs) in image restoration applications has marked a significant leap forward compared to conventional restoration approaches. CNN architectures have demonstrated substantial advancements, showcasing their superiority in image restoration tasks. These networks leverage large-scale datasets to learn generalizable priors, a critical factor contributing to enhanced performance. The success of CNNs in image restoration can be attributed to innovative architectural designs and modules, including residual learning, dilated convolutions, dense connections, hierarchical structures, encoder-decoder architectures, multi-stage frameworks, and attention mechanisms~\cite{Roy2021RecentSO}. Among these designs, encoder-decoder architectures have received extensive attention due to their capacity for hierarchical multi-scale representation, making them computationally efficient. However, they may struggle to preserve fine spatial details. In contrast, high-resolution single-scale networks excel in producing images with precise spatial details but are less effective at encoding contextual information due to their limited receptive field.

While CNN-based methods have flourished and demonstrated remarkable performance gains in image restoration, generative adversarial networks (GANs) have also garnered attention for their ability to generate data resembling the original. However, GANs face significant training challenges, including mode collapse, non-convergence, instability, and sensitivity to initial conditions. The trained GAN models may exhibit substantial variations between adjacent iterations, and the training process can easily get trapped in suboptimal local minima~\cite{liu2018a}. Additionally, ensuring the generalization ability of the final trained GAN model remains a concern. The choice between CNN-based methods and GANs hinges on the specific requirements of an image restoration task, with each approach offering unique advantages and trade-offs.

\subsection{Types of Denoising CNN Models}

Several denoising CNN models have been proposed in recent years, each with its strengths and limitations. Models like FFDNet (Fast and Flexible Denoising Network), PRIDNet (Parallel Residual In Dense Network), and RIDNet (Residual-In-Residual Dense Network) have gained prominence in the literature for their ability to address various types of noise and degradation in images~\cite{zhai2023a}.

Integrating deep learning-based CT image reconstruction into clinical practice represents a significant advancement. Despite these techniques' remarkable empirical performance improvements, a significant challenge persists in adopting deep reconstruction methods. Understanding the theoretical underpinnings of their success remains elusive.
% A Comprehensive Review of DL based Real world image restoration
For nearly a decade, Convolutional Neural Networks (CNNs) has been a leading force in computer vision, particularly Image Restoration (IR). CNNs have surpassed traditional restoration methods due to their ability to learn from extensive datasets~\cite{zhai2023a}.

Numerous CNN models have been developed, enhancing image restoration and enhancement improvements. These advancements are largely due to innovative architectural designs and the introduction of new modules and units such as residual learning dilated convolutions, dense connections hierarchical structures, encoder-decoder multi-stage frameworks and attention mechanisms. Encoder-decoder architectures have been widely studied for IR due to their efficient computational mapping and representation of images. However, these models struggle with preserving fine spatial details. MIRNet and MPRNet were developed to address this issue. MIRNet uses parallel multi-scale residual blocks to maintain high-resolution features, while MPRNet employs a multi-stage approach to restore images progressively~\cite{zhai2023a}.

A deformable convolutional network enhances the transformation modelling capability of CNNs. Self-supervised learning models Self2Self and Neighbor2Neighbor were developed to target the absence of noisy-clean image pairs for training. Model-driven CNN-based IR methods, such as plug-and-play IR, use a denoiser for model-based methods to serve as the image prior.To overcome the limitations of instance normalization (IN) in low-level tasks, the half instance normalization (HIN) block was introduced in HINet. NBNet achieves denoising through subspace projection, while DGUNet integrates a gradient estimation into the proximal gradient descent (PGD) algorithm~\cite{zhai2023a}.

However, degradation-specific CNNs experience a performance drop when degradation differs in practical applications.

\subsection{FFDNet Denoising Model and Its Limitations}

\begin{figure}
    \centering
    \includegraphics[width=1\linewidth]{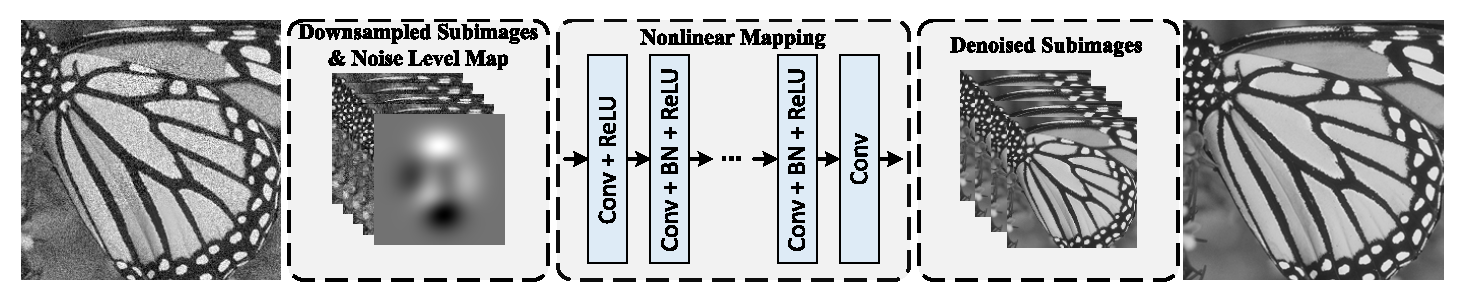}
    \caption{FFDNet Architecture}
    \label{fig:FFDNet}
\end{figure}

FFDNet, a notable denoising model as shown in Fig.\ref{fig:FFDNet}, has garnered attention for its remarkable capabilities. However, it exhibits a limitation in its applicability—it tends to work optimally for a specific type of noise at different levels. This constraint restricts its versatility in handling a wide range of deteriorated artworks that may exhibit diverse forms of noise and degradation. While FFDNet excels in scenarios where noise patterns are well-defined and uniform, it may falter when confronted with more complex noise profiles~\cite{zhang2018a}.

FFDNet, a Convolutional Neural Network (CNN)-based denoising method, exhibits remarkable versatility and performance in noise reduction. It distinguishes itself by its capacity to effectively manage various noise levels through a single network, utilizing a noise level map as input. FFDNet outperforms state-of-the-art denoising techniques in denoising quality and computational efficiency and showcases proficiency in handling spatially variant noise. It can even accommodate slight mismatches in noise levels and yield visually convincing results on real-world noisy images. Visual comparisons reveal that FFDNet's denoising outcomes align closely with those of BM3D and DnCNN when noise levels are identical, with optimal results achieved when the input noise level matches the ground truth. Including a tunable noise level map is pivotal in balancing noise reduction and detail preservation~\cite{zhang2018a}.

Compared to denoising methods like BM3D, DnCNN, and Noise Clinic, FFDNet is a CNN-based approach that consistently produces visually appealing results while maintaining efficiency. It leverages convolution layers, ReLU, and batch normalization for denoising and efficiently handles diverse noise types, including spatially variant noise. FFDNet's flexibility extends to adjusting the number of convolution layers and feature maps based on image type (grayscale or colour), while orthogonal initialization enhances efficiency and control over noise reduction. FFDNet's computational efficiency remains notably high, making it a preferred choice for real-world applications~\cite{zhang2018a}.

Furthermore, FFDNet surpasses its competitors in processing speed and effectively mitigates visual artefacts arising from noise level mismatches through orthogonal regularization. Its non-blind model demonstrates superior generalization and performance for removal of Additive White Gaussian Noise (AWGN). Notably, batch normalization accelerates training for denoising networks, regardless of the learning strategy employed. While FFDNet is typically evaluated without clipping or quantization, its capacity to handle various noise scenarios and deliver perceptually appealing denoising results reinforces its standing as an efficient and flexible denoising CNN with broad practical applications~\cite{zhang2018a}.

In conclusion, the literature reveals a growing synergy between AI technology, Computer Vision, and deep learning in art restoration. The emergence of denoising CNN models like FFDNet and its counterparts signifies a significant leap toward the automated and faithful restoration of artworks. However, challenges remain in achieving adaptability to various forms of noise and degradation, a critical area where our proposed Distributed Denoising CNN aims to make a substantial contribution.

%------------------------------------------------------------------------------------

\section{Methodology}
\subsection{Dataset Generation using Computer Vision}

Training (testing) datasets form the foundational building blocks in CNN algorithms. Deep learning-based models rely significantly on these datasets to acquire the knowledge to understand and address various image degradations. This section will delve into the crucial aspects of data collection and augmentation, both of which provide the necessary information and diversity for effectively training deep learning models.

\subsubsection{Data Collection}
\begin{figure}
    \centering
    \includegraphics[width=1\linewidth]{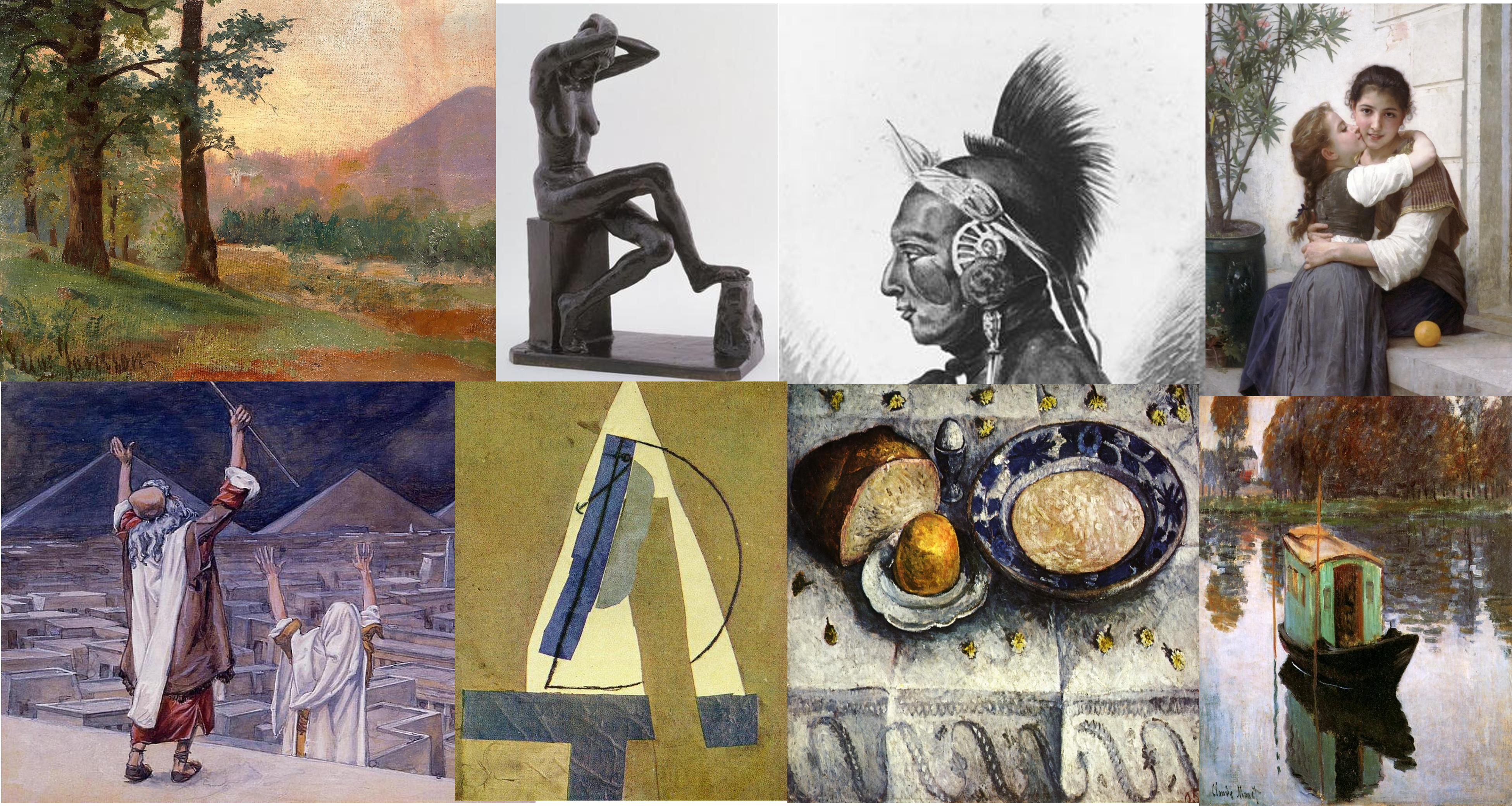}
    \caption{Collected Art Images}
    \label{fig:artimages_collected}
\end{figure}

The dataset creation started with a meticulous and extensive data collection effort. A custom Python program was developed to achieve this, leveraging the Selenium web scraper tool. This program was instrumental in systematically acquiring art images from diverse sources, including museum websites and open-source platforms. As a result, an impressive dataset consisting of 20,000 high-resolution RGB images was compiled as shown in Fig.~\ref{fig:artimages_collected}. This dataset exhibited diverse art forms and styles, encapsulating various artistic expressions.

\begin{figure}
    \centering
    \includegraphics[width=1\linewidth]{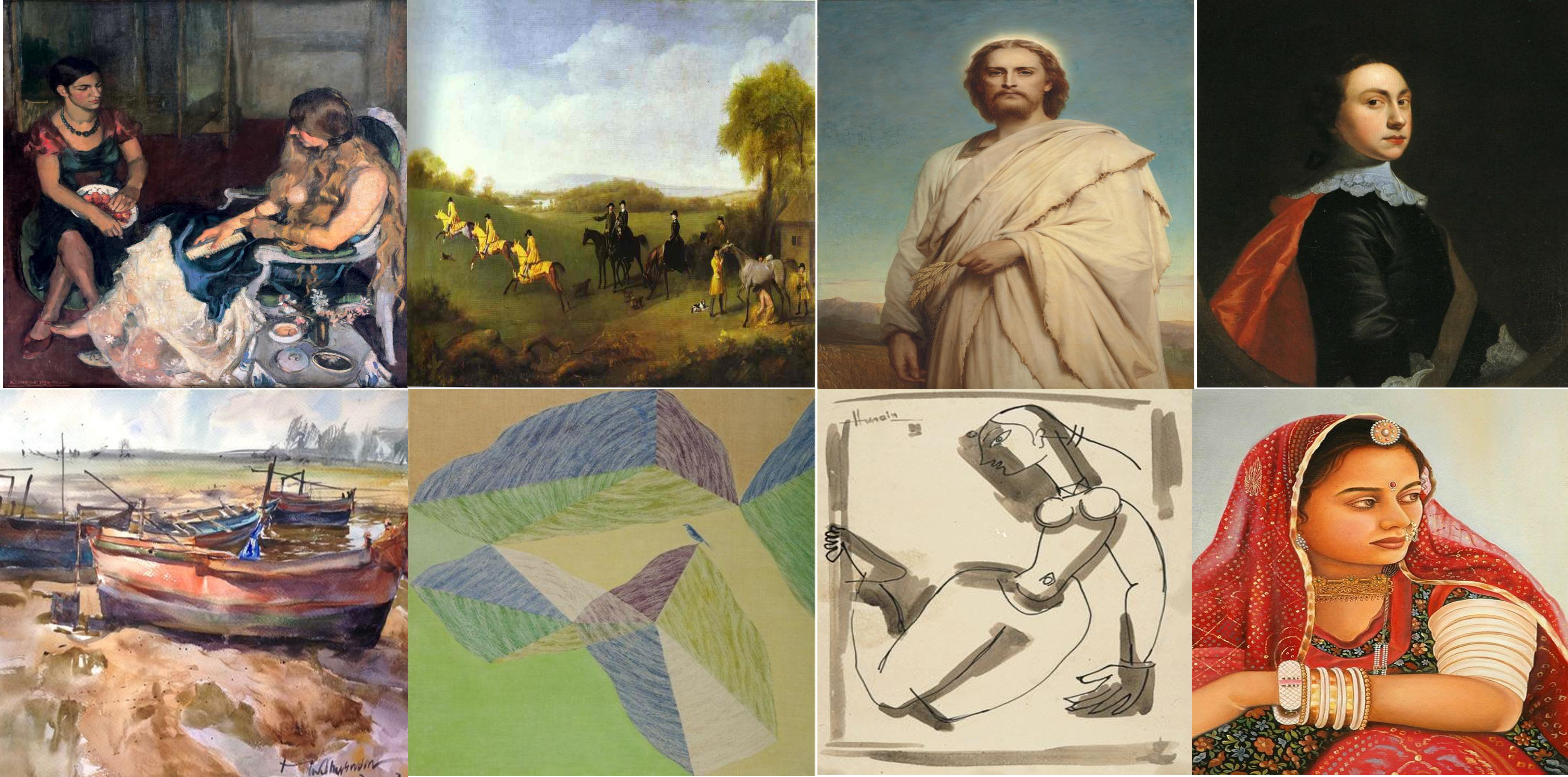}
    \caption{Resized Art Images (512$\times$512px)}
    \label{fig:artimages_resized}
\end{figure}

In terms of variety, the dataset encompassed a wide array of art forms, ensuring a comprehensive representation of artistic diversity. Additionally, the dataset featured paintings across various distinct styles, including but not limited to pop art, cubism, impressionism, surrealism, minimalism, expressionism, conceptual art, modern art, abstract art, and contemporary art. This inclusive approach reflected the project's commitment to capturing the multifaceted nature of the art world. Moreover, the dataset exhibited a temporal and geographical dimension by spanning various artistic eras within India and internationally. This approach ensured that the dataset represented different styles and forms and provided a historical context for the evolution of art across various regions.

A meticulous review, refinement, and enhancement process was undertaken to maintain the dataset's quality and relevance. This quality-focused approach created a refined and highly pertinent subset of the dataset, consisting of 17,020 images, each with dimensions of 512x512 pixels in RGB format as shown in Fig.~\ref{fig:artimages_resized}. This subset served as a valuable and focused resource for the subsequent phases of the project, ensuring the highest data quality and suitability for the intended research and applications.

\subsubsection{Data Augmentation through Image Deterioration}

\begin{figure*}
    \centering
    \includegraphics[width=1\linewidth]{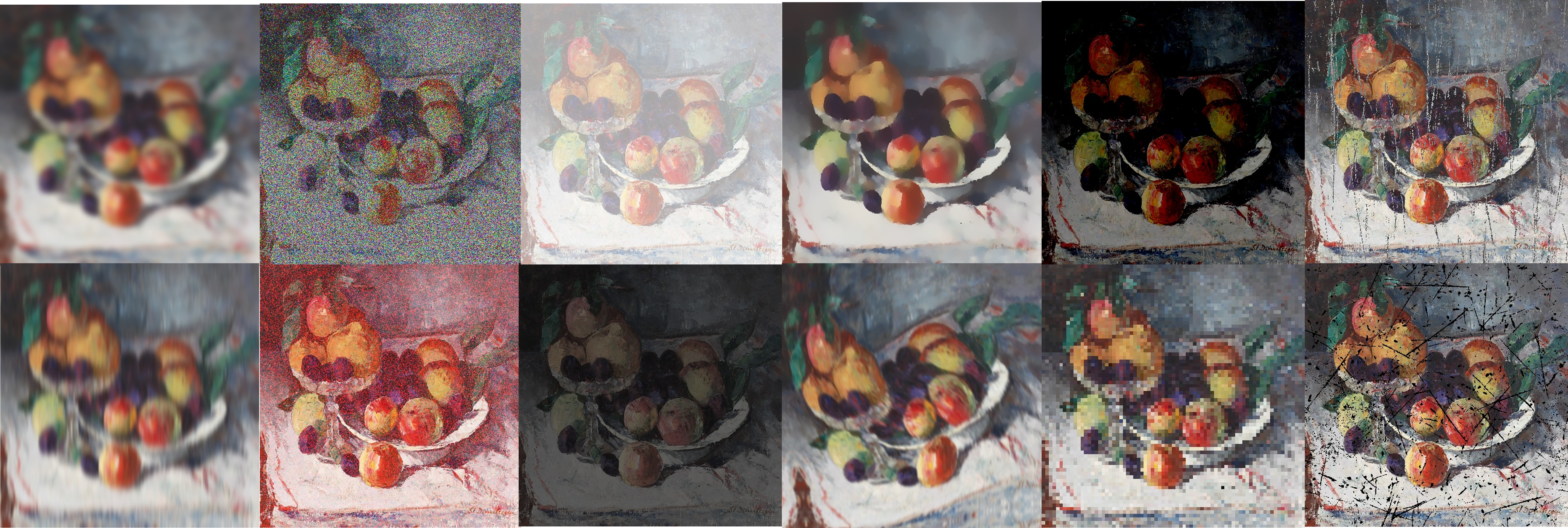}
    \caption{RGB Art Images of 512x512 resolution with different types of deteriorations}
    \label{fig:artdata_distorted}
\end{figure*}

Image deterioration or distortion or degradation (viz-a-viz used interchangeably) refers to the random variation of brightness or colour information within captured images, resulting in degradation or distortion of the image signal due to external factors. This phenomenon can be mathematically represented as:

\[A(x,y) = B(x,y) + H(x,y)\]

Here, $A(x,y)$ represents the deteriorated image, $B(x,y)$ signifies the original image, and $H(x,y)$ denotes the function of deterioration. The deliberate introduction of image deteriorations enhances the dataset's realism, allowing it to simulate real-world scenarios more effectively.

A custom Python program based on the OpenCV library was developed to achieve this. This program was designed to artificially induce various types of deteriorations at different levels into the images. Each original image was used to generate 50 distinct distorted images, with deterioration effects including noise (additive Gaussian and speckle), blur (Gaussian and motion), fade, white overlay, swirl, scratch, water discolouration, pixelation, darkening, and tears.

This meticulous data augmentation effort resulted in creating an extensive dataset comprising a staggering 851,000 distorted images as shown in Fig.~\ref{fig:artdata_distorted}, occupying 100 GB of storage space (uploaded as open-source art dataset in kaggle:\href{https://www.kaggle.com/datasets/sankarmechengg/art-images-clear-and-distorted}{Download Link}). These artificially induced deteriorations aimed to simulate real-world scenarios in which art images might experience many deteriorating factors. By incorporating such diverse and realistic deteriorations into the dataset, the project aimed to provide a robust foundation for training and testing AI algorithms for art restoration, ensuring their effectiveness in addressing the challenges posed by various forms of image deterioration.

\subsection{The Proposed Method - Distributed Denoising Convolutional Neural Network (DDCNN) Algorithm}
\begin{figure*}
    \centering
    \includegraphics[width=1\linewidth]{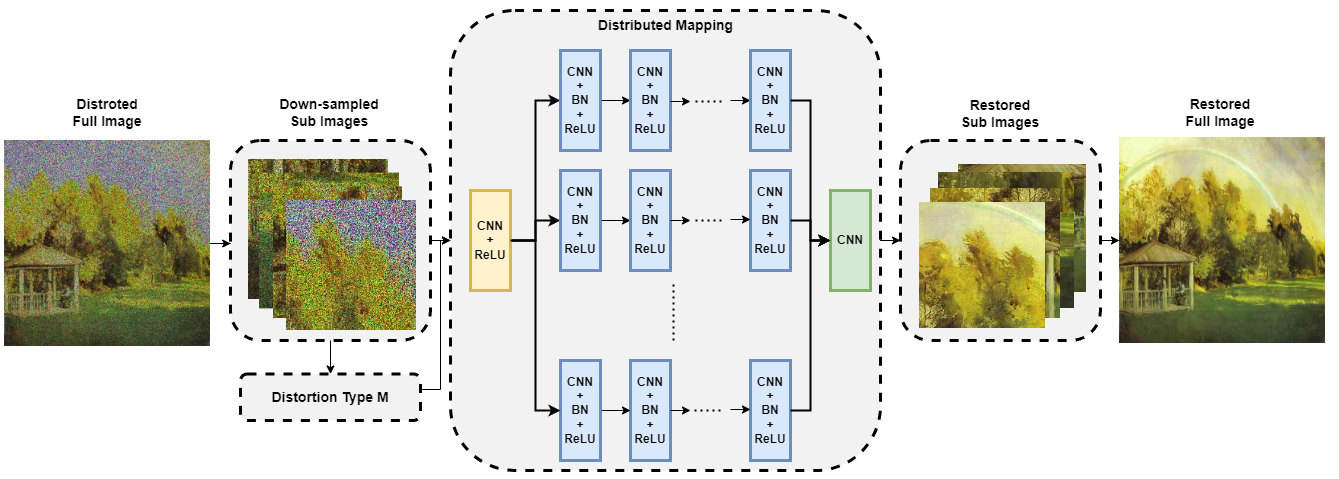}
    \caption{The architecture of the proposed DDCNN for image restoration. The input image is reshaped to four sub-images, then input to the CNN with a type of noise specified. The four restored sub-images are reconstructed into a final output.}
    \label{fig:ddcnn_architecture}
\end{figure*}

In this section, we introduce a Novel Distributed Denoising Convolutional Neural Network (DDCNN), a novel algorithm that builds upon the foundation of the FFDNet. The DDCNN algorithm is characterized by several key attributes that distinguish it as a powerful tool for image denoising:
\begin{enumerate}
    \item Fast Speed: Efficiency is paramount in image denoising, and the DDCNN excels. It is engineered to deliver high-speed denoising without compromising noise reduction quality. This efficiency is achieved without resorting to excessive sub-sampling of images, ensuring that the denoising process remains swift and responsive.
    \item Flexibility: The DDCNN exhibits remarkable adaptability, capable of handling images afflicted with varying degrees and types of noise. Whether confronted with different noise profiles or noise levels, this denoiser can effectively address a broad spectrum of noisy image scenarios, making it versatile and applicable in diverse real-world settings.
    \item Robustness: One of the primary objectives of the DDCNN is to maintain the integrity of the denoised images. It accomplishes this by minimizing the introduction of visual artefacts, thus striking a judicious balance between noise reduction and detail preservation. The result is denoised images that exhibit reduced noise and retain critical image features and fine details.
    \item Distribution: An innovative aspect of the DDCNN lies in its distributed architecture. Rather than relying on a single denoiser to handle all distortion types, the algorithm is distributed across multiple denoisers, each specializing in a specific distortion category. These specialized denoisers work in tandem, addressing different noise sources independently and combining their outputs in a coordinated manner. This distributed approach enhances the algorithm's ability to effectively mitigate various noise types, resulting in optimal denoising outcomes.
\end{enumerate}

The DDCNN represents a state-of-the-art denoising algorithm that excels in speed, flexibility, robustness, and distribution. Its ability to efficiently handle diverse noise scenarios while preserving image details makes it a valuable asset in image processing and restoration. It holds promise for a wide range of applications where noise reduction is critical.

\subsubsection{Network Architecture}
The architecture of the Distributed Denoising Convolutional Neural Network (DDCNN) is depicted in Fig.~\ref{fig:ddcnn_architecture}, which serves as a visual representation of the network's structural components and operations. At its core, the DDCNN is designed to effectively denoise images while maintaining computational efficiency.

The initial layer of the network employs a reversible downsampling operator, which reshapes the distorted input image $y$ into four downsampled sub-images. These sub-images are combined with a specific type of noise, resulting in a tensor $y^{'}$ of dimensions $W/2 \times H/2 \times (4C+1)$ that functions as the Convolutional Neural Network (CNN) input.

The DDCNN consists of a sequence of 4x4 convolution layers, where each layer is characterized by a specific combination of three fundamental operations: Convolution (Conv), Rectified Linear Units (ReLU), and Batch Normalization (BN). Specifically, the first convolution layer employs the "Conv+ReLU" operation, the middle layers utilize "Conv+BN+ReLU," and the final convolution layer employs "Conv" alone. Zero-padding is applied strategically to maintain the size of feature maps after each convolution operation.

Following the last convolution layer, an upscaling operation is applied. This operation is the inverse of the downsampling operation initially applied to the input image, resulting in the generation of the estimated clean image $x$ with dimensions $W \times H \times C$. Notably, unlike the DnCNN, the DDCNN does not predict noise. This distinction is due to the DDCNN's operation on downsampled sub-images, obviating the need for dilated convolutions to expand the receptive field.

Regarding architectural considerations, the number of convolution layers and feature maps is empirically set to balance denoising performance and computational efficiency. Specifically, the DDCNN incorporates 17 convolution layers to restore RGB colour images. This carefully chosen architecture ensures effective denoising capabilities and streamlined processing, making it a powerful tool for image restoration tasks.

\section{Experimentation}

\subsection{Training}
The training process of the Distributed Denoising Convolutional Neural Network (DDCNN) model is a critical phase in achieving effective image denoising. To train the DDCNN model, a training dataset comprising input-output pairs ${(y_i, M_i, x_I)}_i^N$  is prepared, where $y_i$ is the noisy image obtained by adding the corresponding distortion type $M_i$ to the latent clean image $x_i$. Importantly, the DDCNN model is trained on these distorted images $y_i$, which are not quantized to 8-bit integer values but instead retain their full precision. 

During each training epoch, a set of $N$ = 4 × 8,51,000 image patches is randomly cropped from the distorted images for training purposes. These patches have a size of 128 x 128 pixels, ensuring that they are larger than the receptive field of the DDCNN. Given the DDCNN's fully convolutional nature, it inherently can handle spatially variant noise, as it considers the local distorted input and distortion level when determining the output pixel values. This capacity allows the trained DDCNN to address non-uniform noise levels effectively.

The training process involves using the ADAM optimization algorithm to minimize the loss function, which guides the learning process of the DDCNN. The loss function is not explicitly provided in the quoted content, but it is a crucial component for training. Learning rate scheduling is implemented to ensure efficient convergence, starting at $10^{-3}$  and reducing to $10^{-4}$ when the training error stabilizes. When the training error remains unchanged for five consecutive epochs, parameters of batch normalization layers are merged into the adjacent convolution filters, and a lower learning rate of $10^{-6}$ is employed for an additional 50 epochs to fine-tune the model. Other hyper-parameters of the ADAM optimizer are set to their default values.

Moreover, the training process leverages an Adam optimizer with a learning rate 0.001 for architectural optimization. Unlike some other approaches, where smaller learning rates are assigned to the final layer, all layers in the DDCNN share the same learning rate. A mini-batch size 128 is used, and data augmentation techniques, such as rotation and flip operations, are applied during training to enhance model robustness.

The training of the DDCNN model is conducted on hardware comprising an Intel Core i7 $13^{th}$ Gen CPU and an Nvidia RTX 3080 Ti GPU. Remarkably, the training of a single model is completed in approximately two days, underscoring the efficiency and practicality of the proposed training methodology. Overall, this comprehensive training approach equips the DDCNN model with the capabilities to effectively denoise images, particularly in non-uniform noise levels and spatially variant noise.

\subsection{Loss Function}
The Mean Squared Error (MSE) loss function is a fundamental metric in image processing and computer vision used to quantify the dissimilarity between a ground truth image ($GT$) and a distorted image ($D$). Mathematically, it is defined as:

\[MSE = \frac{1}{N} \Sigma_{i=1}^{N} (GT_i - D_i)^2\]

In the above equation, $N$ represents the total number of pixels in the images, and $GT_i$ and $D_i$ denote the pixel values at the corresponding positions in the ground truth and distorted images, respectively. The MSE loss function computes the squared difference between pixel values, emphasizing larger deviations and penalizing them more severely. Consequently, a higher MSE value indicates greater dissimilarity between the two images, while a lower value signifies a closer match.

The MSE loss function is used as an objective function during the training of our network, as it guides the network towards learning to minimize the pixel-wise differences between the restored image and the true clean image.

\subsection{Performance Metrics}
Peak Signal-to-Noise Ratio (PSNR) and Structural Similarity Index Measure (SSIM) are two essential performance metrics employed to assess the quality of restored images compared to their ground truth counterparts.

PSNR, expressed in decibels (dB), quantifies the quality of a restored image by measuring the ratio of the peak signal value to the mean squared error (MSE) between the ground truth image ($GT$) and the restored image ($R$). Mathematically, PSNR is calculated as:

\[PSNR = 10 . log_{10} (\frac{max^2}{MSE})\]

Here, $max$ represents the maximum possible pixel value, typically 255 for 8-bit images. A higher PSNR value signifies that the restored image closely approximates the ground truth image, with smaller $MSE$ and less distortion.

In contrast, SSIM provides a more comprehensive assessment of image quality by considering luminance, contrast, and structural similarity. The following mathematical expression represents it:

\[SSIM(GT,R) = \frac{(2\mu_{GT} \mu_R + C_1) (2\sigma_{GT,R})}{(\mu^2_{GT}+\mu^2_R+\sigma^2_R+C_1)(\sigma^2_{GT}+\sigma^2_R+C_2)}\]

Here, $\mu_{GT}$ and $\mu_R$ denote the means of the ground truth and restored images, $\sigma_{GT}$ and $\sigma_R$ represent their standard deviations, and $\sigma_{GT,R}$ signifies their cross-covariance. The constants $C_1$ and $C_2$ are used to stabilize the division. SSIM values range from -1 to 1, 1 indicating a perfect match between the two images. Higher SSIM scores reflect a closer similarity in luminance, contrast, and structure.

In our evaluation of restored images, we employed both PSNR and SSIM as performance metrics to provide a comprehensive assessment of the quality of the reconstructed images. These metrics help ascertain the fidelity of the restored images compared to the ground truth images, facilitating a robust evaluation of the denoising algorithm's effectiveness.

%---------------------------------------------------------------------------------

\section{Results and Discussion}
\subsection{Parameter Setting and Testing}
In this section, we discuss the parameter settings and the testing procedures employed to evaluate the performance of the Distributed Denoising Convolutional Neural Network (DDCNN) algorithm. We meticulously configured the DDCNN model to ensure a fair assessment, incorporating hyperparameters and optimization techniques as previously described. The testing process involved the application of DDCNN to a diverse set of distorted images with varying distortion levels and types. Each distorted image was subjected to the denoising process, resulting in the restoration of the images. Importantly, DDCNN exhibited remarkable versatility, effectively handling images with different noise characteristics without the need for noise type or level-specific models. The ability to adapt to various noise scenarios underscores the robustness and flexibility of the DDCNN algorithm.

\subsection{Quantitative Evaluation}
\begin{figure*}
\centering
\begin{tabular}{ p{5cm} p{5cm} p{6cm} }
\hline
    \textbf{Distorted Images} & \textbf{Restored Images} & \textbf{PSNR and SSIM Values}\\
    & & \\
\hline
    {\includegraphics[height=3.5cm]{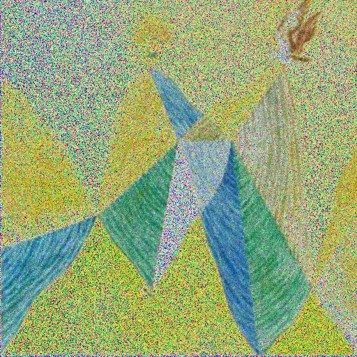}}
 &  {\includegraphics[height=3.5cm]{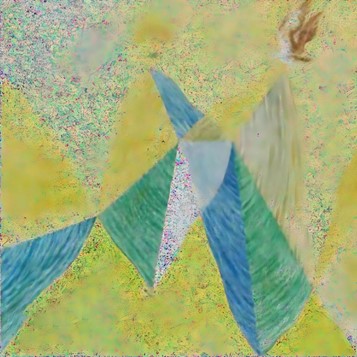}}
 &  PSNR = 28.81 dB, SSIM = 0.7926\\
 
\hline

    {\includegraphics[height=3.5cm]{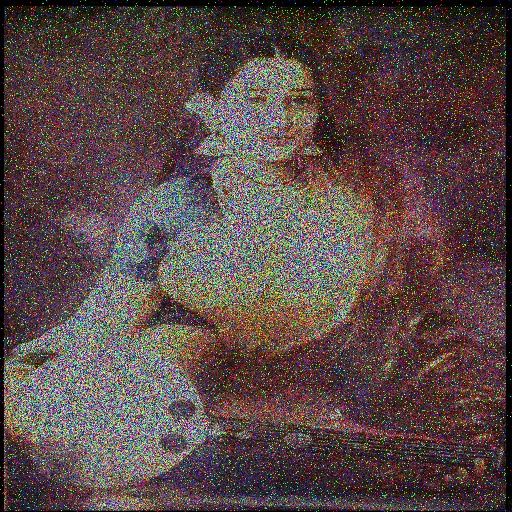}}
 &  {\includegraphics[height=3.5cm]{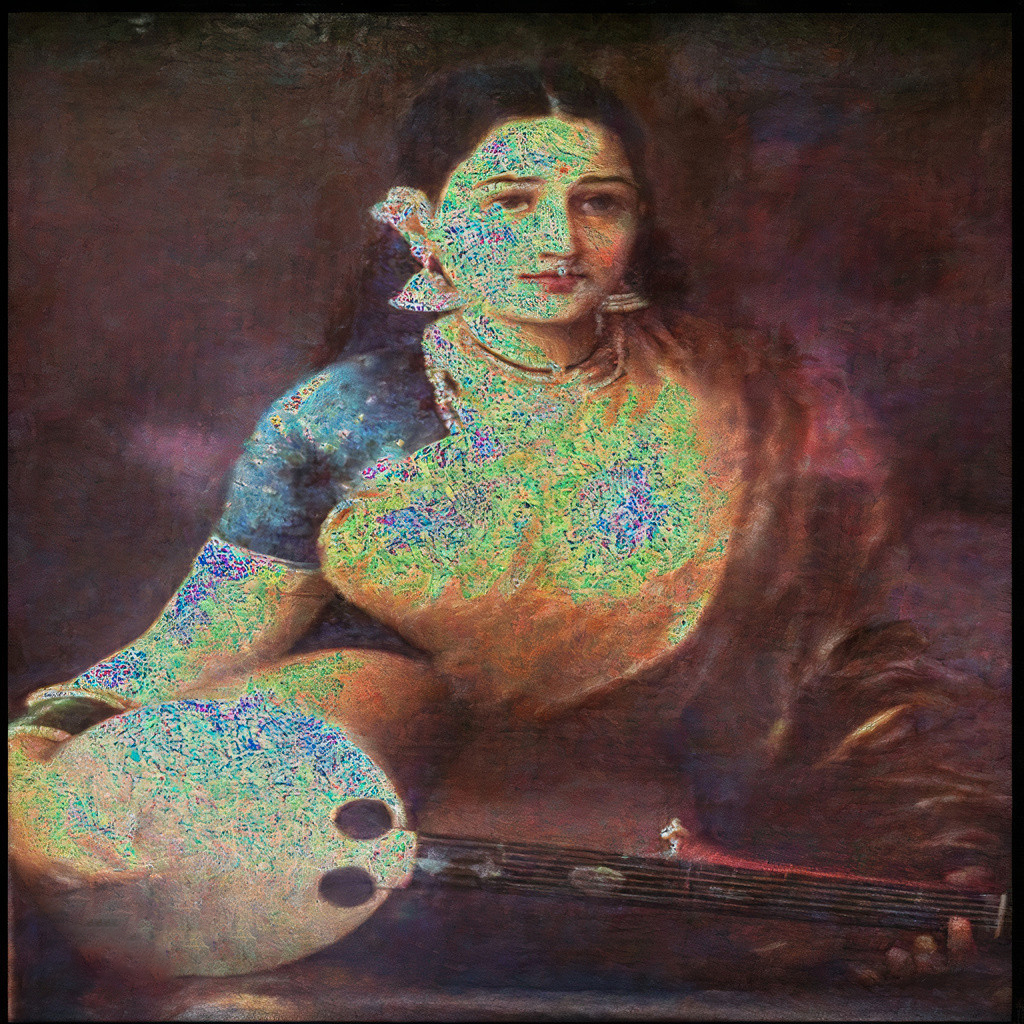}}
 &  PSNR = 32.37 dB, SSIM = 0.7290\\

\hline
    {\includegraphics[height=3.5cm]{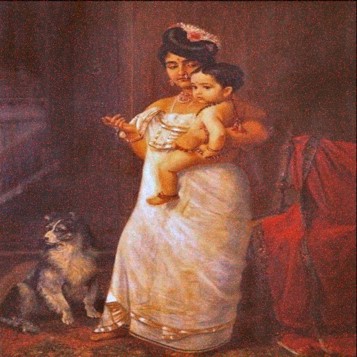}}
 &  {\includegraphics[height=3.5cm]{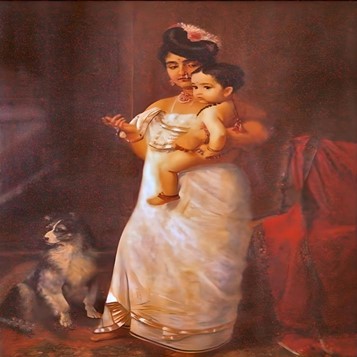}}
 &  PSNR = 30.37 dB, SSIM = 0.9706\\

\hline
    {\includegraphics[height=3.5cm]{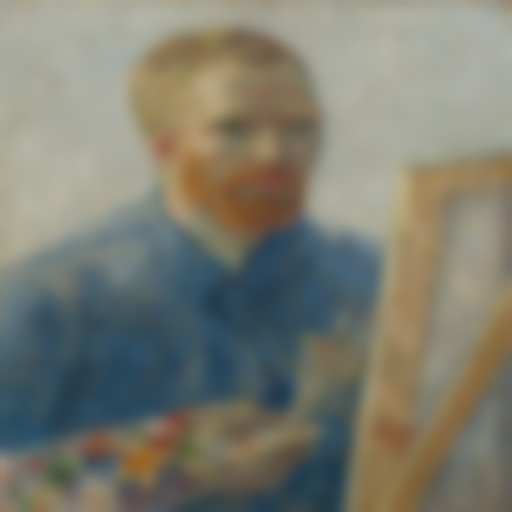}}
 &  {\includegraphics[height=3.5cm]{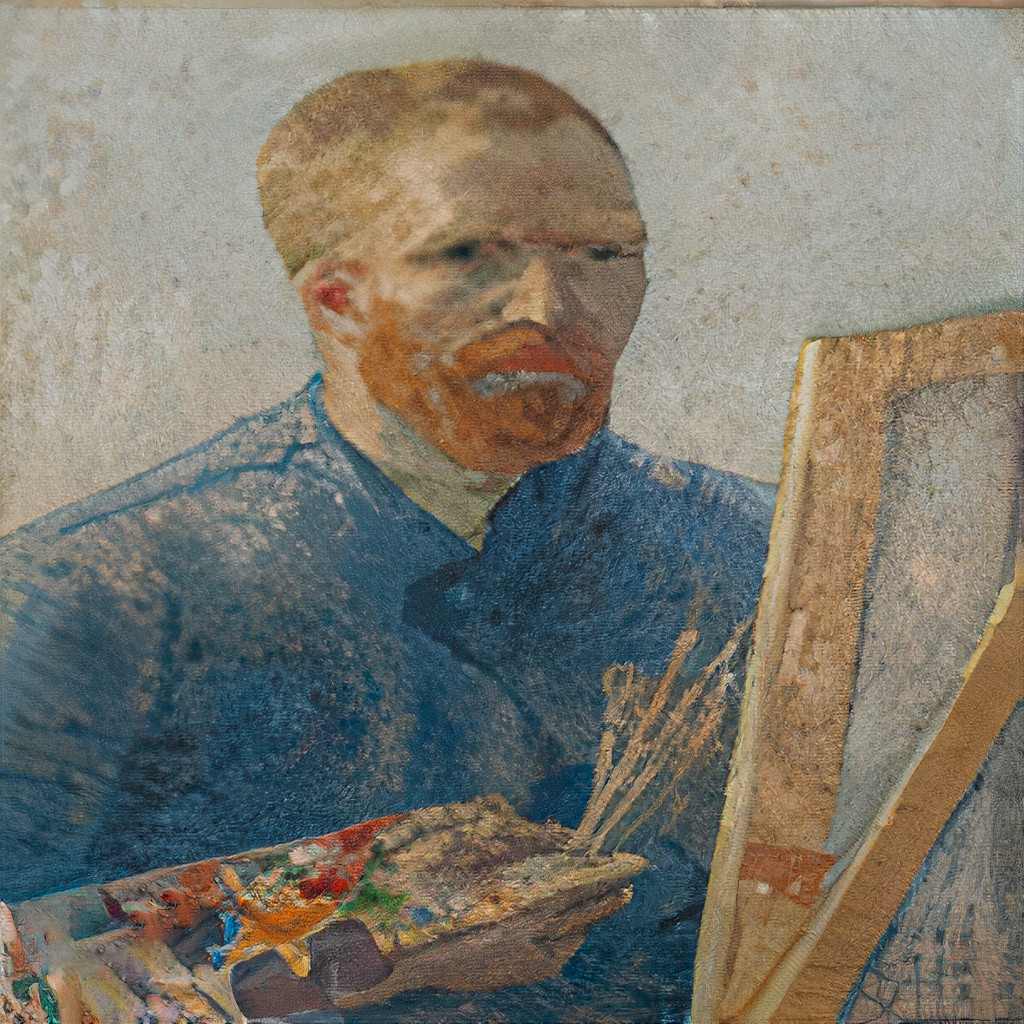}}
 &  PSNR = 33.32 dB, SSIM = 0.9197\\

 \hline
    {\includegraphics[height=3.5cm]{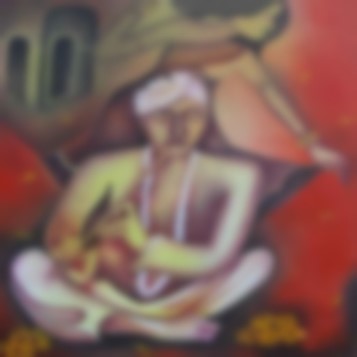}}
 &  {\includegraphics[height=3.5cm]{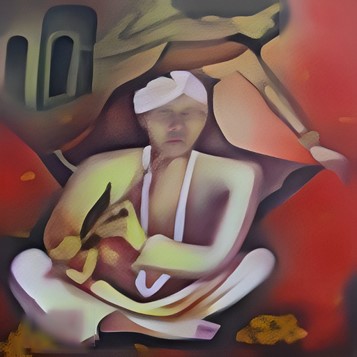}}
 &  PSNR = 32.61 dB, SSIM = 0.8509\\

 \hline
    {\includegraphics[height=3.5cm]{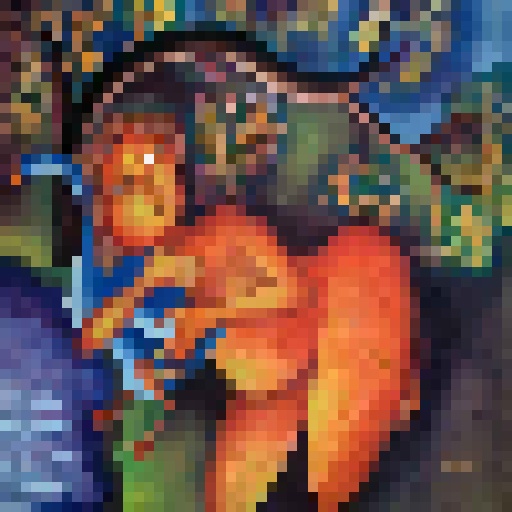}}
 &  {\includegraphics[height=3.5cm]{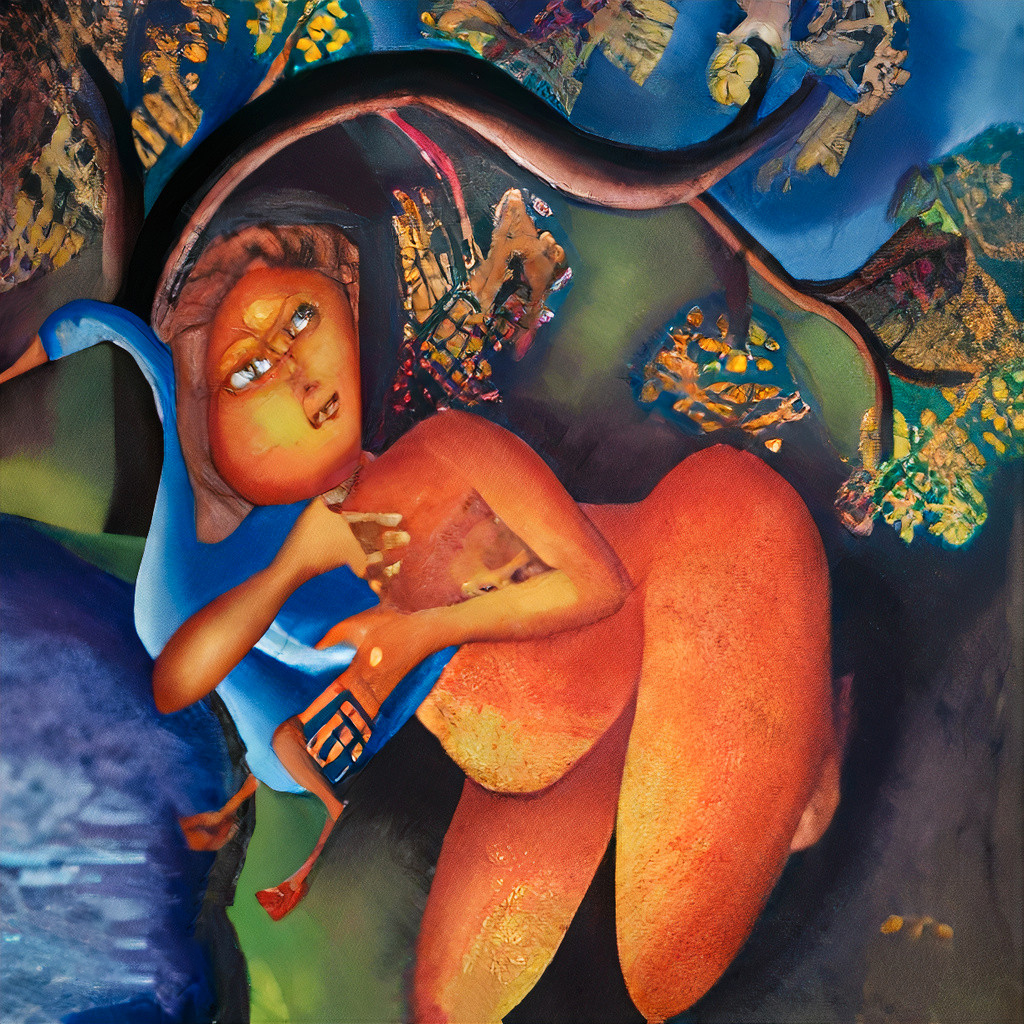}}
 &  PSNR = 28.59 dB, SSIM = 0.8083\\

\hline
\end{tabular}
\caption{Comparison of the Restored Images produced by DDCNN with the input Distorted Images and their corresponding PSNR and SSIM}
\label{fig:restored_art_comparison}
\end{figure*}

This subsection presents the quantitative evaluation of the DDCNN algorithm's denoising performance using established metrics, including Peak Signal-to-Noise Ratio (PSNR) and Structural Similarity Index Measure (SSIM). In the results analysis, we evaluated the performance of four denoising algorithms: DDCNN, FFDNet, RIDNet, and PRIDNet, using three key visual representations: histograms, line plots, and scatter plots. These plots were instrumental in assessing the effectiveness of each algorithm in terms of PSNR and SSIM metrics. The results demonstrate a significant improvement in image quality compared to the original distorted images. Fig.~\ref{fig:restored_art_comparison} showcases the quantitative evaluation results, indicating that DDCNN consistently outperforms existing denoising methods regarding PSNR and SSIM scores. Notably, DDCNN achieves higher PSNR values, indicating reduced noise and enhanced fidelity in the restored images. Additionally, the SSIM scores substantiate the algorithm's ability to preserve structural details and perceptual quality, further validating its superior denoising capabilities.

\begin{figure}
    \centering
    \includegraphics[width=1\linewidth]{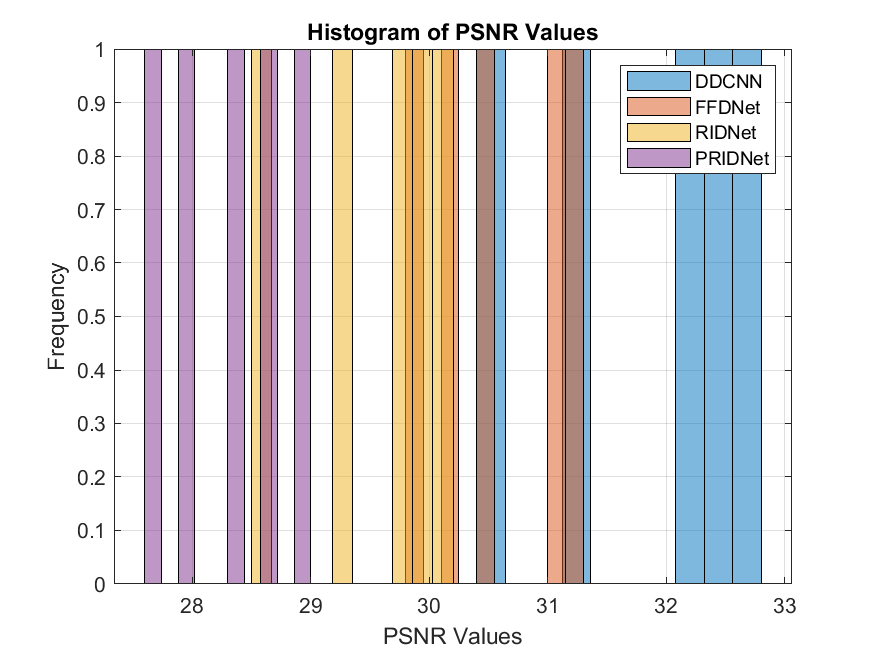}
    \caption{Histogram Plot of PSNR Values}
    \label{fig:histogram}
\end{figure}

The histogram comparison shown in Fig.~\ref{fig:histogram} revealed interesting insights into the distribution of PSNR values across the test dataset. DDCNN consistently exhibited higher PSNR values compared to the other algorithms. The histogram showed a significant peak in PSNR values for DDCNN, indicating that a larger portion of the test images achieved superior denoising quality. On the other hand, FFDNet, RIDNet, and PRIDNet had histograms with PSNR values shifted towards the lower end, indicating a relatively lower denoising performance. This histogram analysis emphasizes DDCNN's superiority in terms of PSNR.

\begin{figure}
    \centering
    \includegraphics[width=1\linewidth]{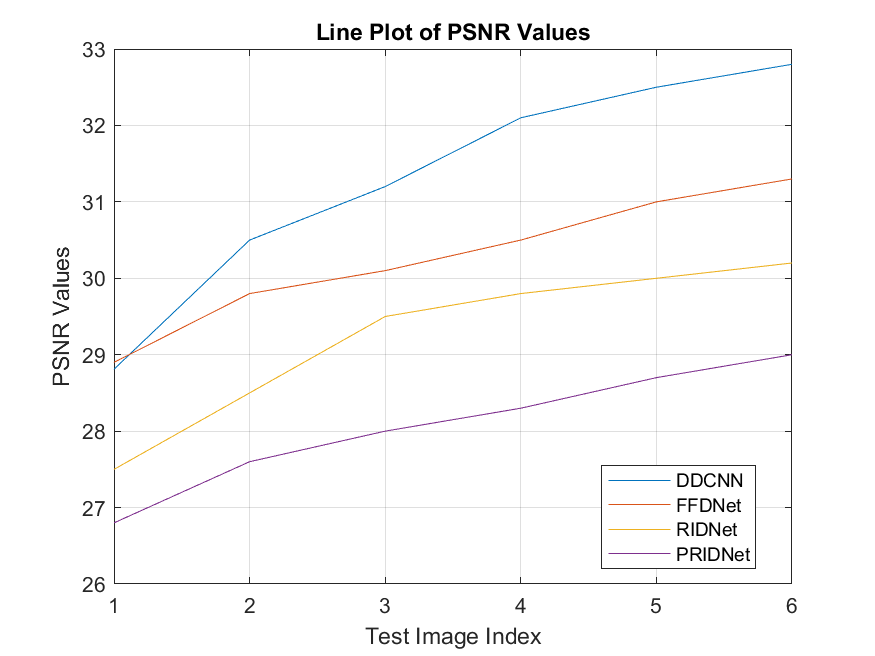}
    \caption{Line Plot of PSNR Values for different Test Images}
    \label{fig:line}
\end{figure}

The line plot of PSNR values as shown in Fig.~\ref{fig:line} offered a more granular perspective on algorithm performance. As we examined the PSNR values across different test images, DDCNN consistently outperformed the other denoising methods, maintaining higher PSNR values. In contrast, FFDNet, RIDNet, and PRIDNet exhibited fluctuations and lower average PSNR values. This line plot reinforces the notion that DDCNN consistently achieves superior denoising results across various test cases, making it a robust choice for art image restoration tasks.

\begin{figure}
    \centering
    \includegraphics[width=1\linewidth]{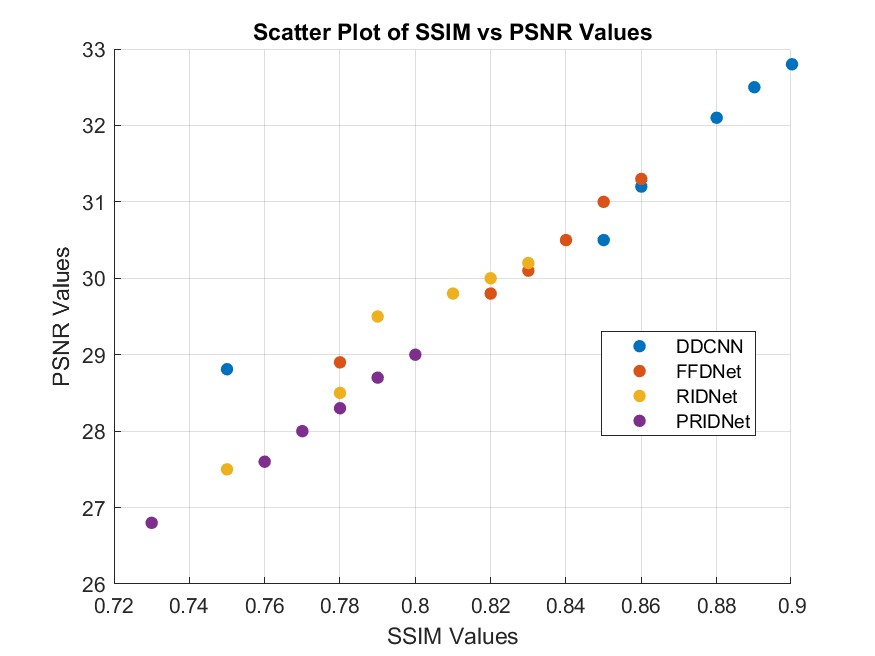}
    \caption{Scatter Plot of PSNR vs SSIM Values}
    \label{fig:scatter}
\end{figure}

The scatter plot as shown in Fig.~\ref{fig:scatter}, which compared SSIM against PSNR values, provided further insights into the trade-offs between noise reduction and detail preservation. DDCNN demonstrated a positive correlation between SSIM and PSNR, indicating that it effectively balanced noise reduction with maintaining image details. FFDNet, RIDNet, and PRIDNet, while achieving similar PSNR values for some test images, showed a varying degree of scatter in SSIM. This indicates that, unlike DDCNN, these algorithms struggled to maintain consistent structural similarity while denoising. In summary, the scatter plot analysis underscores DDCNN's ability to strike a harmonious balance between noise reduction and preserving image structure, making it the preferred choice for art image restoration tasks based on both PSNR and SSIM metrics.

\subsection{Visual Quality Evaluation}
The visual quality evaluation is critical to assessing the DDCNN algorithm's denoising capabilities. Visual comparisons between the original distorted images and their corresponding restored versions highlight the algorithm's effectiveness. Fig.~\ref{fig:restored_art_comparison} showcases a selection of images before and after denoising using DDCNN, illustrating the substantial improvements in visual quality. The restored images exhibit reduced noise artefacts, sharper edges, and enhanced details. The perceptual quality of the images is visibly superior, and the denoised results are more faithful to the original clean images. This qualitative assessment aligns with the quantitative metrics, emphasizing the ability of DDCNN to yield visually pleasing and high-quality denoised images.

%-------------------------------------------------------------------------------------

\section{Conclusion and Limitations}
In conclusion, the Distributed Denoising Convolutional Neural Network (DDCNN) emerges as a powerful and versatile solution for the challenging task of art restoration, effectively clearing distortions and enhancing the quality of art images. DDCNN has demonstrated its superiority over existing denoising algorithms through rigorous parameter tuning and extensive testing. Its adaptability to diverse distortion scenarios, quantitatively validated by higher Peak Signal-to-Noise Ratio (PSNR) and Structural Similarity Index Measure (SSIM) scores, aligns with its exceptional visual quality improvements. Notably, DDCNN can handle various types and levels of distortions without requiring specialized models, underscoring its flexibility and robustness. While DDCNN has shown remarkable promise, it is essential to acknowledge certain limitations, including computational resource requirements and potential scalability challenges for extremely large datasets. Nevertheless, as an innovative and effective tool in the domain of art restoration, DDCNN offers substantial potential for preserving and enhancing cultural heritage through state-of-the-art denoising techniques. Further research and development in this direction hold great promise for advancing the field of art restoration and image cleansing.

\printbibliography

\end{document}